\RequirePackage{fix-cm}
\documentclass[smallextended]{svjour3}       
\smartqed  
\usepackage{cite}
\usepackage{amsmath,amssymb,amsfonts}
\usepackage{graphicx}
\usepackage{textcomp}
\usepackage{url}
\usepackage{multirow}
\usepackage{paralist}
\usepackage{algorithm}
\usepackage{algorithmic}
\usepackage{color}
\usepackage{subcaption}
\usepackage{natbib}
\usepackage{bm}
\newtheorem{mydef}{Definition}
\newtheorem{myprob}{Problem}

\usepackage{amssymb,mathtools}
\begin{document}

\title{ResGCN: Attention-based Deep Residual Modeling for Anomaly Detection on Attributed Networks}

\titlerunning{ResGCN}        

\author{Yulong Pei         \and
        Tianjin Huang \and \\
        Werner van Ipenburg \and \\
        Mykola Pechenizkiy
}


\institute{Y. Pei\and T. Huang \and M. Pechenizkiy \at
              Department of Mathematics and Computer Science \\
              Eindhoven University of Technology, 5600 MB Eindhoven, the Netherlands\\
              \email{\{y.pei.1,t.huang,m.pechenizkiy\}@tue.nl}           
            \and
            W. van Ipenburg \at
            Cooperatieve Rabobank U.A., the Netherlands\\
            \email{werner.van.ipenburg@rabobank.nl}
}

\date{Received: date / Accepted: date}

\maketitle

\begin{abstract}
Effectively detecting anomalous nodes in attributed networks is crucial for the success of many real-world applications such as fraud and intrusion detection. 
Existing approaches have difficulties with three major issues: sparsity and nonlinearity capturing, residual modeling, and network smoothing. We propose Residual Graph Convolutional Network (ResGCN), an attention-based deep residual modeling approach that can tackle these issues: modelling the attributed networks with GCN allows to capture the sparsity and nonlinearity;  utilizing a deep neural network allows to directly learn residual from the input, and a residual-based attention mechanism reduces the adverse effect from anomalous nodes and prevents over-smoothing. 
Extensive experiments on several real-world attributed networks demonstrate the effectiveness of ResGCN in detecting anomalies.
\end{abstract}

\section{Introduction}
\label{intro}
Attributed networks are ubiquitous in a variety of real-world applications. Data from many real-world domains can be represented as attributed networks, where nodes represent entities with attributes and edges express the interactions or relationships between entities. Different from plain networks where only structural information exists, attributed networks also contain rich features to provide more details to describe individual elements of the networks. For instance, in social networks, user profiles contain important information to describe users. In citation networks, paper abstracts can provide complementary information to the citation structures. In gene regulatory networks, gene sequence expressions are the attributes beside the interactions between molecular regulators. Due to the ubiquity of attributed networks, various data mining tasks on attributed networks have attracted an upsurge of interest such as community detection~\citep{falih2018community,li2018community,pei2015nonnegative}, link prediction~\citep{barbieri2014follow,li2018streaming,brochier2019link}, network embedding~\citep{huang2017label,huang2017accelerated,meng2019co}, etc.

\begin{figure}
	\begin{center}
		\includegraphics[width=0.7\textwidth]{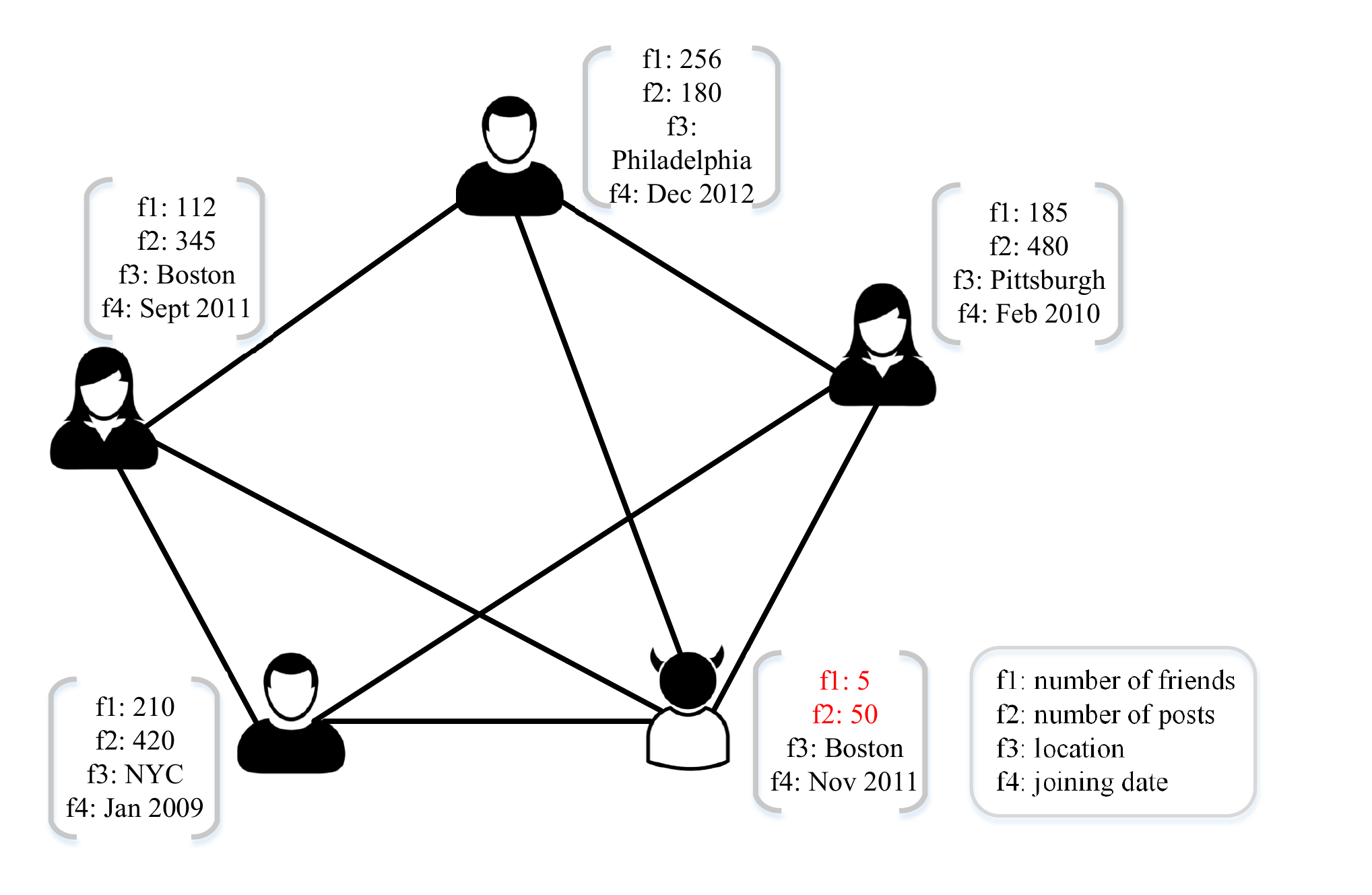}
		\caption{An illustration of failure in previous message passing based anomaly detection approaches.}
		\label{fig:example}
	\end{center}
\end{figure}

Anomaly detection is one of the most vital problems among these tasks on attributed networks because of its significant implications in a wide range of real-world applications including cyber attack detection in computer networks, fraud detection in finance and spammers discovery in social media, to name a few. 
It is more challenging to detect anomalies on attributed networks because both attributes and structures should be taken into consideration in order to detect anomalous nodes. An illustration is shown in Figure~\ref{fig:example}. The anomalous node is different from others because: 1)~structurally it connects to all other nodes and 2)~its attributes are significantly different from the majority.

Several approaches for anomaly detection on attributed networks have been proposed recently in the literature. Most of them aim at detecting anomalies in an unsupervised fashion because of the prohibitive cost for accessing the ground-truth anomalies~\citep{ding2019deep}. They can be categorized into four types: community analysis, subspace selection, residual analysis and deep learning methods. Community analysis methods~\citep{gao2010community} detect anomalies by identifying the abnormality of current node with other nodes within the same community. Subspace selection approaches~\citep{perozzi2014focused} first learn a subspace for features and then discover anomalies in that learned subspace. Residual analysis methods~\citep{li2017radar,peng2018anomalous} explicitly model the residual information by reconstructing the input attributed network based on matrix factorization. Deep learning methods use deep neural networks to capture the nonlinearity of networks and detect anomalies in an unsupervised~\citep{ding2019deep} or supervised way~\citep{liang2018semi}.

However, there are three major issues in existing approaches: sparsity and nonlinearity capturing, residual modeling, and network smoothing. \textit{Capturing sparsity and nonlinearity} is important in anomaly detection on networks because real-world attributed networks are complex and non-linear. Previous shallow models such as non-negative matrix factorization~\citep{li2017radar,peng2018anomalous} fail to detect anomalies because of the incapability of modeling nonlinearity.  Although \textit{residual modeling} has been explored in previous studies~\citep{li2017radar,peng2018anomalous}, the residual information has been modeled from the reconstruction error. Thus, they cannot be adaptively learned from the input networks. \textit{Smoothing networks}, which is based on the homophily hypothesis~\citep{mcpherson2001birds}, is a commonly used strategy to detect anomalies on networks, e.g.,~\citep{ding2019deep}. However, these methods are not in line with anomaly detection because they might over-smooth the node representations, and make anomalous nodes less distinguishable from the majority~\citep{li2019specae}.

To tackle these issues, in this paper, we propose Residual Graph Convolutional Network (ResGCN), a novel approach for anomaly detection on attributed networks. ResGCN is capable of solving the above three problems as follows: (1) to \textit{capture the sparsity and nonlinearity} of networks, ResGCN is based on GCN to model the attributed networks; (2) to \textit{model residual information}, ResGCN learns residual directly from the input using a deep neural network; and (3) to \textit{prevent over-smoothing of node representations}, ResGCN incorporates the attention mechanism based on learned residual information. Thus, the information propagation of anomalous nodes can be reduced. 
The contributions of this paper are summarized as follows:
\begin{itemize}
    \item We propose novel anomaly detection method named ResGCN. ResGCN captures the sparsity and nonlinearity of networks using GCN, learns the residual information using a deep neural network, and reduces the adverse effect from anomalous nodes using the residual-based attention mechanism.
    \item We propose a residual information based anomaly ranking strategy and the residual information is learned from the input network instead of reconstruction errors.
    \item We conduct extensive experiments on real-world attributed networks. Experimental results demonstrate the effectiveness of our proposed ResGCN in the task of anomaly detection w.r.t. different evaluation metrics.
\end{itemize}

The rest of this paper is organized as follows. Section~\ref{prob} formally defines the problem of anomaly detection on attributed networks. Section~\ref{model} introduces the proposed ResGCN model for anomaly detection. Section~\ref{exp} provides empirical evidence of ResGCN performance on anomaly detection in real-world networks w.r.t. different evaluation metrics. Section~\ref{rw} briefly discusses related work on anomaly detection on attributed networks. Finally, we conclude in Section~\ref{conc}.

\section{Problem Definition}
\label{prob}
We first summarize some notations and definitions used in this papers. Following the commonly used notations, we use bold uppercase characters for matrices, e.g., $\bm{X}$, bold lowercase characters for vectors, e.g., $\bm{b}$, and normal lowercase characters for scalars, e.g., $c$. The The $i^{th}$ row of a matrix $X$ is denoted by $X_{i,:}$ and $(i,j)^{th}$ element of matrix $X$ is denoted as $X_{i,j}$. The Frobenius norm of a matrix is represented as $\|\cdot\|_F$ and $\|\cdot\|_2$ is the $L_2$ norm. In detail, the main symbols are listed in Table~\ref{tb:notation}.

\begin{table}
\centering
\caption{Table of notations.}
\label{tb:notation}
\begin{tabular}{c|l}
\hline
   Symbol       &  Description   \\ \hline
$V$       & node set   \\ \hline
$E$      &  edge set   \\ \hline
$m$ & number of edges  \\ \hline
$n$  &  number of nodes\\ \hline
$d$  &  number of attributes\\ \hline
$\bm{A}$    &  adjacency matrix \\ \hline
$\bm{X}$     &   attribute matrix  \\ \hline
$\bm{W}^{l}$   & the trainable weight matrix in the $l^{th}$ layer    \\ \hline
$\bm{H}^{l}$   & the latent representation matrix in the $l^{th}$ layer     \\ \hline
$\bm{R}^{l}$   & the residual matrix  in the $l^{th}$ layer     \\ \hline
$\alpha$  & the trade-off parameter for reconstruction error  \\ \hline
$\lambda$  & the residual parameter  \\ \hline
\end{tabular}
\end{table}

\begin{mydef}
\textbf{Attributed Networks}. An attributed network $\mathcal{G} = \{V,E,\bm{X}\}$ consists of: (1) a set of nodes $V=\{v_1,v_2,...,v_n\}$, where $|V|=n$ is the number of nodes; (2) a set of edges $E$, where $|E|=m$ is the number of edges; and (3) the node attribute matrix $\bm{X}\in \mathbb{R}^{n\times d}$, the $i^{th}$ row vector $\bm{X}_{i,:}\in\mathbb{R}^{d},~i = 1,...,n$ is the attribute of node $v_i$.
\end{mydef}

The topological structure of attributed network $\mathcal{G}$ can be represented by an adjacency matrix $\bm{A}$, where $A_{i,j}=1$ if there is an edge between node $v_i$ and node $v_j$. Otherwise, $A_{i,j}=0$. We focus on the undirected networks in this study and it is trivial to extend it to directed networks. The attribute of $\mathcal{G}$ can be represented by an attribute matrix $\bm{X}$. Thus, the attributed network can be represented as $\mathcal{G} = \{\bm{A},\bm{X}\}$. With these notations and definitions, same to previous studies~\citep{li2017radar,peng2018anomalous,ding2019deep}, we formulate the task of anomaly detection on attributed networks:
\begin{myprob}
\textbf{Anomaly Detection on Attributed Networks}. Given an attributed network $\mathcal{G} = \{\bm{A},\bm{X}\}$, which is represented by the adjacency matrix $\bm{A}$ and attribute matrix $\bm{X}$, the task of anomaly detection is to find a set of nodes that are rare and differ singularly from the majority reference nodes of the input network.
\end{myprob}

\section{Proposed Method}
\label{model}
In this section we first introduce the background of GCN. Next, we present the proposed model ResGCN in details. Then we analyze the complexity of ResGCN.

\begin{figure*}
	\begin{center}
		\includegraphics[width=\textwidth]{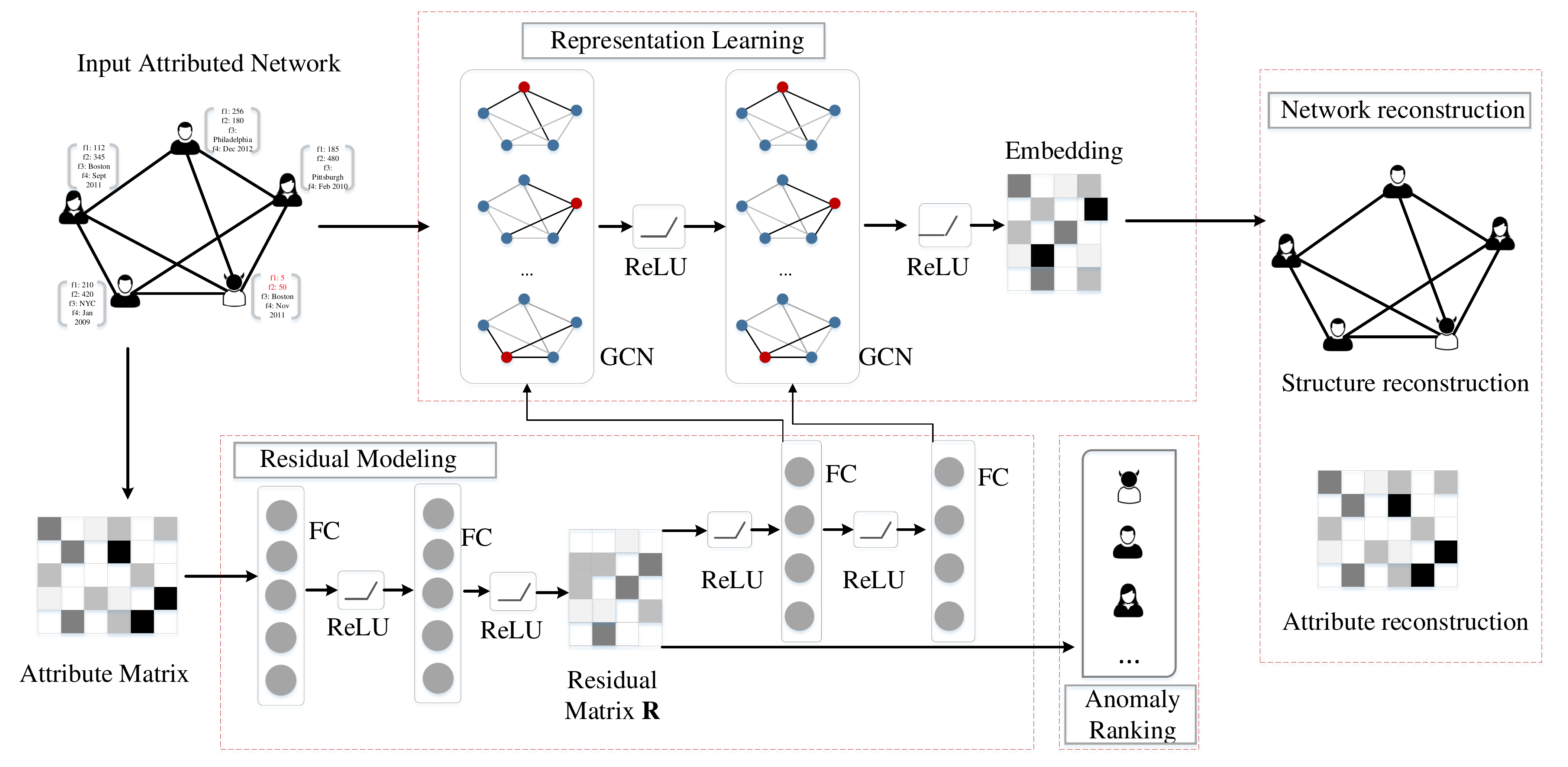}
		\caption{The framework of our proposed ResGCN.}
		\label{fig:frame}
	\end{center}
\end{figure*}

\subsection{Graph Convolutional Networks}
GCN learns node representations by passing and aggregating messages between neighboring nodes. Different types of GCN have been proposed recently~\citep{kipf2016semi,hamilton2017inductive}, and we focus on one of the most widely used versions proposed in~\citep{kipf2016semi}. Formally, a GCN layer is defined as
\begin{equation}
\label{gcn}
    \bm{h}^{(l+1)}_i=f\Big(\sum_{j\in Ne(i)}\frac{1}{\sqrt{\widetilde{\bm{D}}_{i,i}\widetilde{\bm{D}}_{j,j}}}\bm{h}^{(l)}_j \bm{W}^{(l)}\Big),
\end{equation}
where $\bm{h}^{(l)}_i$ is the latent representation of node $v_i$ in layer $l$, $Ne(i)$ is the set of neighbors of node $v_i$, and $\bm{W}^{l}$ is the layer-specific trainable weight matrix. $f(\cdot)$ is a non-linear activation function and we select ReLU as the activation function following previous studies~\citep{kipf2016semi} (written as $f_{ReLU}(\cdot)$ below). $\widetilde{\bm{D}}$ is the diagonal degree matrix of $\widetilde{\bm{A}}$ defined as $\widetilde{D}_{i,i}=\sum_{j}\widetilde{A}_{i,j}$ where $\widetilde{\bm{A}}=\bm{A}+\bm{I}$ is the adjacency matrix of the input attributed network $\bm{G}$ with self connections $\bm{I}$. Equivalently, we can rewrite GCN in a matrix form:
\begin{equation}
\label{gcn-mat}
	\bm{H}^{(l+1)}=f_{ReLU}\Big(\widetilde{\bm{D}}^{-\frac{1}{2}}\widetilde{\bm{A}}\widetilde{\bm{D}}^{-\frac{1}{2}}\bm{H}^{(l)}\bm{W}^{(l)}\Big).
\end{equation}
For the first layer, $\bm{H}^{(0)}=\bm{X}$ is the attribute matrix of the input network. Therefore, we have
\begin{equation}
\label{gcn-1st}
	\bm{H}^{(1)}=f_{ReLU}\Big(\widetilde{\bm{A}}\bm{X}\bm{W}^{(0)}\Big).
\end{equation}
The architecture of GCN can be trained end-to-end by incorporating task-specific loss functions. In the original study, GCN aims at semi-supervised classification task so the cross-entropy loss is evaluated by adding the softmax function as the output of the last layer.Formally, the overall cross-entropy error is evaluated on the graph for all the labeled samples:
\begin{equation}
	\mathcal{L}_{cls}=-\sum_{i\in L}\sum_{c=1}^C \bm{Y}_{ic}\log \hat{\bm{Y}}_{ic}
\end{equation}
where $L$ is the set of nodes with labels, $C$ is the number of classes, $\bm{Y}$ is the label and $\hat{\bm{Y}}=softmax(\bm{H})$ is the prediction of GCN passing the hidden representation in the final layer $\bm{H}^{(L)}$ to a softmax function.

Note that original GCN~\citep{kipf2016semi} is designed for semi-supervised learning, our target is to detect anomalies in an unsupervised way. Therefore, the cross entropy loss for (semi-)supervised learning is not suitable in our problem settings. We will introduce our proposed loss function which is based on network reconstruction errors in the following section.

\subsection{ResGCN}
In this section, we present the proposed framework of ResGCN in details. ResGCN consists of four components: residual modeling, representation learning, network reconstruction and anomaly ranking. The architecture of this model is illustrated in Figure~\ref{fig:frame}.

\subsubsection{Residual Modeling}
Although some previous studies explicitly model the residual information for anomaly detection on attributed networks, e.g., Radar~\citep{li2017radar} and ANOMALOUS~\citep{peng2018anomalous}, these methods have two major limitations: (1) They are based on linear models, e.g., matrix factorization, so these shallow models are incapable of capturing the nonlinearity of networks.
(2) The residual information has been modeled from the reconstruction error. Thus, they cannot be adaptively learned from the input networks. However, real-world networks are complex and residual information has different patterns in different datasets. Motivated by the study~\citep{dabkowski2017real}, which proposes to learn the saliency map based on convolutional network, we propose to use a deep neural network to learn the residual by capturing the nonlinearity in ResGCN. Formally, 
\begin{equation}
\label{fc}
	\bm{R}^{(l+1)}=f_{ReLU}(\bm{R}^{(l)}\cdot \bm{W}^{(l)}),
\end{equation}
where $\bm{R}^{(l)}$ is the input for the fully connected (FC) layer $l$, and $\bm{W}^{(l)}$ is the layer-specific trainable weight matrix which needs to be learned during the training of the model. The output of this network is the residual matrix, denoted as $\bm{R}$. 

Another aim of the residual modeling component is to learn the attention weights to control the message passing in network representation based on the residual information. Similarly, we use FC layer which takes the residual matrix $\bm{R}$ as input and the calculation is the same to Eq (\ref{fc}). Each output of the FC layer corresponds to the attention weights for each GCN layer shown in Figure~\ref{fig:frame}. Therefore, the number of FC layers to learn the weights is equal to the number of GCN layers which will be presented below.

\subsubsection{Representation Learning}
The second component of ResGCN aims at learning representations of the input attributed network. Our proposed representation learning method can not only capture the sparsity and nonlinearity of networks but also prevent the information propagating of anomalies. In this component, we adopt GCN with attention which is based on the residual information modeled in the first component to learn the embeddings of nodes. To make the computations tractable, we follow~\citep{zhu2019robust} and assume all hidden representations of nodes are independent. Therefore, we can aggregate node neighbors as follows: 
\begin{equation}
\label{gcn-ne}
    \bm{h}^{(l)}_{Ne(i)}=\sum_{j\in Ne(i)}\frac{1}{\sqrt{\widetilde{\bm{D}}_{i,i}\widetilde{\bm{D}}_{j,j}}}\bm{h}^{(l)}_j.
\end{equation}

To prevent the information propagation from the anomalous nodes, we propose an attention mechanism based on the residual information modeled by the first component to assign different weights to neighbors. The reason is that it is intuitive the nodes with larger residual errors are more likely to be anomalies~\citep{li2017radar}. Motivated by~\citep{zhu2019robust}, we use the smooth exponential function to control the effect of residual information on weights. Formally, the weight is defined as
\begin{equation}
	\bm{\theta}^{(l)}_j=\exp (-\gamma \bm{R}_{j}^{(l)}),
\end{equation}
where $\bm{\theta}^{(l)}_j$ are the attention weights of node $v_j$ in the $l^{th}$ layer and $\gamma$ is a hyper-parameter. By taking the attention weights into account, the modified aggregated node neighbor representation can be written as:
\begin{equation}
\label{resgcn-ne}
    \bm{h}^{(l)}_{Ne(i)}=\sum_{j\in Ne(i)}\frac{1}{\sqrt{\widetilde{\bm{D}}_{i,i}\widetilde{\bm{D}}_{j,j}}}\bm{h}^{(l)}_j\circ\bm{\theta}^{(l)}_j,
\end{equation}
where $\circ$ is the element-wise product. Then we apply learnable filters and non-linear activation function (ReLU used in this study) to $\bm{h}^{(l)}_{Ne(i)}$ in order to calculate $\bm{h}^{(l)}_i$. Formally the layer is defined as:
\begin{equation}
\label{gcn-obj}
    \bm{h}^{(l+1)}_i=f\Big(\sum_{j\in Ne(i)}\frac{1}{\sqrt{\widetilde{\bm{D}}_{i,i}\widetilde{\bm{D}}_{j,j}}}\Big(\bm{h}^{(l)}_j\circ\bm{\theta}_j\Big)\bm{W}^{(l)}\Big).
\end{equation}
Equivalently, the matrix form is:
\begin{equation}
    \bm{H}^{(l+1)}=f_{ReLU}\Big(\widetilde{\bm{D}}^{-\frac{1}{2}}\widetilde{\bm{A}}\widetilde{\bm{D}}^{-\frac{1}{2}}\Big(\bm{H}^{(l)}\circ\bm{\Theta}\Big)\bm{W}^{(l)}\Big),
\end{equation}
where $\bm{\Theta}=\exp(-\gamma \bm{R}^{(l)})$. Similarly, for the first layer, we have
\begin{equation}
    \bm{H}^{(1)}=f_{ReLU}\Big(\widetilde{\bm{A}}\bm{X}\bm{W}^{(0)}\Big).
\end{equation}
The output of the last GCN layer is the node embedding matrix $\bm{Z}$.

\subsubsection{Network Reconstruction}
The target of the third component of ResGCN is to reconstruct the network which consists of structure reconstruction and attribute reconstruction. Both reconstructions are based on the latent representation $\bm{Z}$ learned in the representation learning component.

\paragraph{Structure Reconstruction}
Let $\hat{\bm{A}}$ denote the reconstructed adjacency matrix. Following~\citep{ding2019deep,kipf2016variational}, we use the inner product of the latent representations between two nodes to predict if an edge exists between them. Intuitively, if the latent representations of two nodes are similar, it is more likely that there is an edge between them. Formally, the prediction between two nodes $v_i$ and $v_j$ can represented as follows:
\begin{equation}
    P(\hat{A}_{i,j}=1|\bm{z}_i,\bm{z}_j)=f_{sigmoid}(\bm{z}_i,\bm{z}_j),
\end{equation}
where $f_{sigmoid}$ function is to convert the prediction as a probability value. Accordingly, the whole reconstructed network structure based on the latent representations $\bm{Z}$ can be represented as follows:
\begin{equation}
    \hat{\bm{A}}=sigmoid(\bm{Z}\bm{Z}^T).
\end{equation}
Correspondingly, the reconstruction error for structure can be represented as:
\begin{equation}
\label{struc_decoder}
  E_S=\|\bm{A}-\hat{\bm{A}}\|_F^2 .
\end{equation}

\paragraph{Attribute Reconstruction} 
To reconstruct the original attributes, DOMINANT~\citep{ding2019deep} uses another graph convolution layer as the decoder to reconstruct the attributes. However, considering that graph convolution is simply a special form of Laplacian smoothing and mixes the nodal features and its nearby neighbors~\citep{li2018deeper}, we adopt the multi-layer perception as our decoder instead. Formally, let $\hat{\bm{X}}$ be the reconstructed attributes and the reconstruction process can be formalized as follows:
\begin{equation}
\label{attri_decoder_sip}
  \hat{\bm{X}}=\Phi^{n}(\bm{Z}),
\end{equation}
where $n$ denotes the number of FC layers and $\Phi^{n}(\cdot)$ denotes n-layer perception which is composed with linear functions followed by non-linear activation function. By taking the residual into consideration, the attribute reconstruction is:
\begin{equation}
\label{attri_decoder}
  E_A=\|\bm{X}-\hat{\bm{X}}-\lambda \bm{R}\|_F^2,
\end{equation}
where $\lambda$ is the residual parameter to control how much residual information we want to use in the attribute reconstruction error. This error is similar to~\citep{li2017radar,peng2018anomalous} which explicitly incorporate the residual information in attribute reconstruction.

Based on the structure and attribute reconstruction errors, we can propose the objective function of our proposed ResGCN model. To jointly learn the reconstruction errors, the objective function of ResGCN is defined as the weighted combination of two errors:
\begin{align}
\label{recons}
    \mathcal{L}&=(1-\alpha)E_S+\alpha E_A\\\nonumber
    &=(1-\alpha)\|\bm{A}-\hat{\bm{A}}\|_F^2+\alpha \|\bm{X}-\hat{\bm{X}}-\lambda \bm{R}\|_F^2,
\end{align}
where $\alpha$ is the trade-off parameter to control the importance of errors from structure and attributed reconstruction. By minimizing the objective function, we aim to approximate the input attributed network based on the latent representations. Different from previous studies which rank reconstruction errors to detect anomalous nodes~\citep{ding2019deep}, in our proposed model, we rank the residual matrix $\bm{R}$ for anomaly identification. Formally, the anomaly score for node $v_i$ is
\begin{equation}
    score(v_i)=\|\bm{R}_{i,:}\|_2.
\end{equation}
Finally, the anomalies are the nodes with larger scores and we can detect anomalies according to the ranking of anomaly scores. This ranking strategy is superior to reconstruction error based methods because in our model the residual is explicitly learn from the data and implicitly updated by minimizing the reconstruction error. Therefore, it can better capture the anomaly of the data and less be adversely influenced by the noise from the model.

\subsection{Complexity Analysis}
The computational complexity of GCN is linear to the number of edges on the network. For a particular layer, the convolution operation is $\widetilde{\bm{D}}^{-\frac{1}{2}}\widetilde{\bm{A}}\widetilde{\bm{D}}^{-\frac{1}{2}}\bm{X}\bm{W}$ and its complexity is $O(edf)$~\citep{ding2019deep}, where $e$ is the number of non-zero elements in the adjacency matrix $\bm{A}$, $d$ is the dimensions of attributes, and $f$ is the number of feature maps of the weight matrix. For network reconstruction, we use link prediction  to reconstruct the structure and multi-layer perception to reconstruct the attribute both of which are pairwise operations. Thus, the overall complexity is $O(ed\bm{F}+n^2)$ where $\bm{F}$ is the summation of all feature maps across different layers.

\section{Experiments}
\label{exp}
In this section, we evaluate the effectiveness of our proposed ResGCN model on several real-world datasets and present experimental results in order to answer the following three research questions.
\begin{itemize}
    \item \textbf{RQ1}: Does ResGCN improve the anomaly detection performance on attributed networks?
    \item \textbf{RQ2}: Is deep residual matrix ranking strategy effective in identifying anomalies?
    \item \textbf{RQ3}: How do the parameters in ResGCN affect the anomaly detection performance?
\end{itemize}

\subsection{Datasets}
In order to evaluate the effectiveness of our proposed method, we conduct experiments on two types of real-world attributed networks: data with and without ground-truth anomaly labels. All networks have been widely used in previous studies~\citep{li2017radar,peng2018anomalous,ding2019deep,gutierrez2019multi}:
\begin{itemize}
    \item Networks with ground-truth anomaly labels: Amazon and Enron\footnote{\url{https://www.ipd.kit.edu/mitarbeiter/muellere/consub/}}. Amazon is a co-purchase network~\citep{muller2013ranking}. It contains 28 attributes for each node describing properties about online items including rating, price, etc. The anomalous nodes are defined as nodes having the tag \textit{amazonfail}. Enron is an email network~\citep{metsis2006spam} where each node is an email with 20 attributes describing metadata of the email including content length, number of recipients, etc, and each edge indicates the email transmission between people. Spammers are labeled as the anomalies in Enron data. The details of these attributed networks are shown in Table~\ref{tb:ad_data}.
    \item Networks without ground-truth anomaly labels: BlogCatalog, Flickr and ACM\footnote{\url{http://people.tamu.edu/~xhuang/Code.html}}. BlogCatalog is a blog sharing website where users are the nodes and following relations between users are edges. Each user is associated with a list of tags to describe themselves and their blogs, which are used as attributes. Flickr is an image hosting and sharing website. Similarly, users and user following relations are nodes and edges, respectively. Tags are the attributes. ACM is a citation network where each node is a paper and each edge indicates a citation relation between papers. Paper abstracts are used as attributes. The details of these attributed networks are shown in Table~\ref{tb:ne_data}.
\end{itemize}
For the networks with labels, we directly use these provided labels to evaluate our method. For the data without labels, we need to manually inject anomalies for empirical evaluation. To make a fair comparison, we follow previous studies for anomaly injection~\citep{ding2019deep}. In specific, two anomaly injection methods have been used to inject anomalies by perturbing topological structure and nodal attributes, respectively:
\begin{itemize}
    \item \textbf{Structural anomalies}: structural anomalies are generated by perturbing the topological structure of the network. It is intuitive that in real-world networks, small cliques are typically anomalous in which a small set of nodes are much more connected to each other than average~\citep{skillicorn2007detecting}. Thus, we follow the method used in~\citep{ding2019deep,ding2019interactive} to generate some small cliques. In details, we randomly select $s$ nodes from the network and then make those nodes fully connected, and then all the $s$ nodes forming the clique are labeled as anomalies. $t$ cliques are generated repeatedly and totally there are $s\times t$ structural anomalies.
    \item \textbf{Attribute anomalies}: we inject an equal number of anomalies from structural perspective and attribute perspective. Same to~\citep{ding2019deep,song2007conditional}, $s\times t$ nodes are randomly selected as the attribute perturbation candidates. For each selected node $v_i$, we randomly select another $k$ nodes from the network and calculate the Euclidean distance between $v_i$ and all the $k$ nodes. Then the node with largest distance is selected as $v_j$ and the attributes $\bm{X}_j$ of node $v_j$ is changed to $\bm{X}_i$ of node $node_i$. The selected node $v_j$ is regarded as the attribute anomaly.
\end{itemize}
In the experiments, we set $s=15$ and set $t$ to 10, 15, and 20 for BlogCatalog, Flickr and ACM, respectively which are the same to~\citep{ding2019deep} in order to make the comparison with DOMINANT~\citep{ding2019deep}. To facilitate the learning process, in our experiments, we follow~\citep{ding2019interactive} to reduce the dimensionality of attributes using Principal Component Analysis (PCA) and the dimension is set to 20.

\begin{table}
\centering
\caption{Statistics of networks with ground-truth anomaly labels.}
\label{tb:ad_data}
\begin{tabular}{l|c|c}
\hline
\multicolumn{1}{c|}{} & Amazon  & Enron \\ \hline
\# nodes                & 1,418 & 13,533    \\ \hline
\# edges                & 3,695 & 176,987    \\ \hline
\# attributes              & 28     & 20          \\ \hline
\# anomalies        & 28 & 5     \\ \hline
\end{tabular}
\end{table}

\begin{table}
\centering
\caption{Statistics of networks without ground-truth anomaly labels.}
\label{tb:ne_data}
\begin{tabular}{l|c|c|c}
\hline
\multicolumn{1}{c|}{} & BlogCatalog  & Flickr & ACM \\ \hline
\# nodes                & 5,196 & 7,575    & 16,484 \\ \hline
\# edges                & 171,743 & 239,738    & 71,980 \\ \hline
\# attributes           & 8,189 & 12,074    & 8,337   \\ \hline
\# anomalies              & 300     & 450        & 600      \\ \hline
\end{tabular}
\end{table}

\subsection{Evaluation Metrics}
In the experiments, we use two evaluation metrics to validate the performance of these anomaly detection approaches:
\begin{itemize}
    \item \textbf{ROC-AUC}: we use the area under the receiver operating characteristic curve (ROC-AUC) as the evaluation metric for anomaly detection as it has been widely used in previous studies~\citep{li2017radar,peng2018anomalous,ding2019deep,gutierrez2019multi}. ROC-AUC can quantify the trade-off between true positive rate (TP) and false positive rate (FP) across different thresholds. The TP is defined as the detection rate, i.e. the rate of true anomalous nodes correctly identified as anomalous, whereas the FP is the false alarm rate, i.e. rate of normal nodes identified as anomalous~\citep{gutierrez2019multi}.
    \item \textbf{Precision@K and Recall@K}: Since we use the ranking strategy to detect anomalies, measures used in ranking-based tasks such as information retrieval and recommender systems can be utilized to evaluate the performance. In specific, we use Precision@K to measure the proportion of true anomalies that an approach discovered in its top K ranked nodes and Recall@K to measure the proportion of true anomalies that a  method discovered in the total number of ground truth anomalies.
\end{itemize}

\subsection{Baselines}
To demonstrate the effectiveness of our proposed framework in detecting anomalies, we compare the proposed ResGCN model with the following anomaly detection methods:
\begin{itemize}
    \item \textbf{LOF}~\citep{breunig2000lof} measures how isolated the object is with respect to the surrounding neighborhood and detects anomalies at the contextual level. LOF only considers nodal attributes.
    \item \textbf{AMEN}~\citep{perozzi2016scalable} uses both attribute and network structure information to detect anomalous neighborhoods. Specifically, it analyzes the abnormality of each node from the ego-network point of view.
    \item \textbf{Radar}~\citep{li2017radar} is an unsupervised anomaly detection framework for attributed networks. It detects anomalies whose behaviors are singularly different from the majority by characterizing the residuals of attribute information and its coherence with network information.
    \item \textbf{ANOMALOUS}~\citep{peng2018anomalous} is a joint anomaly detection framework to optimize attribute selection and anomaly detection using CUR decomposition of matrix and residual analysis on attributed networks.
    \item \textbf{DOMINANT}~\citep{ding2019deep} utilizes GCN to learn a low-dimensional embedding representations of the input attributed network and then reconstruct both the topological structure and nodal attributes with these representations. Anomalies are selected by ranking the reconstruction errors.
    \item \textbf{MADAN}~\citep{gutierrez2019multi} is a multi-scale anomaly detection method. It uses the heat kernel as filtering operator to exploit the link with the Markov stability to find the context for anomalous nodes at all relevant scales of the network.
\end{itemize}
In the experiments, for our proposed ResGCN, we propose to optimize the loss function with Adam~\citep{kingma2014adam} algorithm and train the proposed model for 100 epochs. We set the learning rate to 0.01. For representation learning, we use two GCN layers (64-neuron and 32-neuron), and for residual modeling, we use three FC layers (\# neuron is equal to \# features) to learn the residual matrix, two FC layers (both 64-neuron) to learn the attention weights for the GCN hidden representation and two FC layers (both 32-neuron) to learn the attention weights for the GCN embedding. For these baselines, we use the default parameters used in the original papers.

\subsection{Experimental Results}
We conduct experiments to evaluate the performance of our proposed model ResGCN by comparing it with several baselines on two different types of networks: networks with and without ground-truth anomaly labels. The experimental results w.r.t. ROC-AUC for networks with ground-truth labels are shown in Table~\ref{tb:auc}. It can be observed from these results:
\begin{itemize}
    \item The proposed ResGCN model outperforms other baseline methods on Amazon data and achieves comparable result on Enron data. It demonstrates the effectiveness of ResGCN.
    \item Deep models such as DOMINANT and residual analysis based methods such as Radar and ANOMALOUS are superior to traditional approaches such as LOF and AMEN. It further validates the effectiveness of deep models and residual modeling.
\end{itemize}
The experimental results w.r.t. Precision@K and Recall@K for networks without ground-truth labels are shown from Table~\ref{tb:prek:1} to Table~\ref{tb:reck:3} respectively. From these evaluation results, some conclusions can be drawn:
\begin{itemize}
    \item The proposed ResGCN model outperforms other baseline methods on all three attributed networks except Precision@50 on Flickr. It demonstrates the effectiveness of our method by combining residual modeling and deep representation learning using deep neural networks to detect anomalies.
    \item Superiority of ResGCN to other approaches in Precision@K and Recall@K indicates our proposed model can not only achieve higher detection accuracy but also find more true anomalies within the ranking list of limited length.
    \item Anomaly detection approaches using deep architecture achieve better performance including ResGCN and DOMINANT. It verifies the importance of nonlinearity modeling for anomaly detection on attributed networks.
    \item The residual analysis based models, i.e., Radar and ANOMALOUS, although fail in capturing the nonlineariry of networks, achieve better performance than conventional approaches such as LOF. It demonstrates the rationality of explicit residual modeling in anomaly detection.
\end{itemize}

\begin{table}
\centering
\caption{Performance of different anomaly detection methods w.r.t. ROC-AUC. The bold indicates the best performance of all the methods.}
\label{tb:auc}
\begin{tabular}{l|c|c}
\hline
          & Amazon & Enron \\ \hline
LOF~\citep{breunig2000lof}       & 0.490  & 0.440 \\ \hline
AMEN~\citep{perozzi2016scalable}      & 0.470  & 0.470 \\ \hline
Radar~\citep{li2017radar}     & 0.580  & 0.650 \\ \hline
ANOMALOUS~\citep{peng2018anomalous} & 0.602   & \textbf{0.695} \\ \hline
DOMINANT~\citep{ding2019deep}  & 0.625   &  0.685   \\ \hline
MADAN~\citep{gutierrez2019multi}     & 0.680  & 0.680 \\ \hline
ResGCN (Our Model)   & \textbf{0.710}  &  0.660       \\ \hline
\end{tabular}
\end{table}

\begin{table*}
\centering
\caption{Performance of different anomaly detection methods w.r.t. precision@K on BlogCatalog. The bold indicates the best performance of all the methods.}
\label{tb:prek:1}
\begin{tabular}{l|c|c|c|c}
\hline
K         & 50    & 100    & 200    & 300     \\ \hline
LOF~\citep{breunig2000lof}       & 0.300 & 0.220  & 0.180  & 0.183  \\ \hline
Radar~\citep{li2017radar}     & 0.660 & 0.670  & 0.550  & 0.416  \\ \hline
ANOMALOUS~\citep{peng2018anomalous} & 0.640 & 0.650  & 0.515  & 0.417  \\ \hline
DOMINANT~\citep{ding2019deep}  & 0.760 & 0.710  & 0.590  & 0.470  \\ \hline
MADAN~\citep{gutierrez2019multi}     & 0.600   & 0.620   & 0.520  & 0.410  \\ \hline
ResGCN (Our Model)   & \textbf{0.848} & \textbf{0.860} & \textbf{0.670} & \textbf{0.483} \\ \hline
\end{tabular}
\end{table*}

\begin{table*}
\centering
\caption{Performance of different anomaly detection methods w.r.t. precision@K on Flickr. The bold indicates the best performance of all the methods.}
\label{tb:prek:2}
\begin{tabular}{l|c|c|c|c}
\hline
K         & 50    & 100    & 200    & 300    \\ \hline
LOF~\citep{breunig2000lof}        & 0.420& 0.380 & 0.270 & 0.237\\ \hline
Radar~\citep{li2017radar}      & 0.740& 0.700 & 0.635 & 0.503\\ \hline
ANOMALOUS~\citep{peng2018anomalous}  & \textbf{0.790}& 0.710 & 0.650 & 0.510\\ \hline
DOMINANT~\citep{ding2019deep}    & 0.770& 0.730 & 0.685& 0.593\\ \hline
MADAN~\citep{gutierrez2019multi}   & 0.710 & 0.680 & 0.620  & 0.540  \\ \hline
ResGCN (Our Model)    & 0.780 & \textbf{0.830} & \textbf{0.875} & \textbf{0.680} \\ \hline
\end{tabular}
\end{table*}

\begin{table*}
\centering
\caption{Performance of different anomaly detection methods w.r.t. precision@K on ACM. The bold indicates the best performance of all the methods.}
\label{tb:prek:3}
\begin{tabular}{l|c|c|c|c}
\hline
K         & 50    & 100    & 200    & 300    \\ \hline
LOF~\citep{breunig2000lof}       &0.060&0.060 &0.045 & 0.037\\ \hline
Radar~\citep{li2017radar}     &0.560&0.580 &0.520 & 0.430\\ \hline
ANOMALOUS~\citep{peng2018anomalous} &0.600&0.570 &0.510 & 0.410\\ \hline
DOMINANT~\citep{ding2019deep}  &0.620&0.590 &0.540 & 0.497\\ \hline
MADAN~\citep{gutierrez2019multi}    & 0.580 & 0.540     & 0.560  & 0.420  \\ \hline
ResGCN (Our Model)   & \textbf{0.812} & \textbf{0.780} & \textbf{0.675} & \textbf{0.573}\\ \hline
\end{tabular}
\end{table*}


\begin{table*}
\centering
\caption{Performance of different anomaly detection methods w.r.t. recall@K on BlogCatelog. The bold indicates the best performance of all the methods.}
\label{tb:reck:1}
\begin{tabular}{l|c|c|c|c}
\hline
K         & 50    & 100    & 200    & 300     \\ \hline
LOF~\citep{breunig2000lof}        & 0.050 & 0.073  & 0.120  & 0.183 \\ \hline
Radar~\citep{li2017radar}     & 0.110 & 0.223  & 0.367  & 0.416 \\ \hline
ANOMALOUS~\citep{peng2018anomalous} & 0.107 & 0.217  & 0.343  & 0.417  \\ \hline
DOMINANT~\citep{ding2019deep}  & 0.127 & 0.237  & 0.393  & 0.470 \\ \hline
MADAN~\citep{gutierrez2019multi}  & 0.105 & 0.215  & 0.375  & 0.380 \\ \hline
ResGCN (Our Model)  & \textbf{0.143} & \textbf{0.299} & \textbf{0.456} & \textbf{0.483} \\ \hline
\end{tabular}
\end{table*}

\begin{table*}
\centering
\caption{Performance of different anomaly detection methods w.r.t. recall@K on Flickr. The bold indicates the best performance of all the methods.}
\label{tb:reck:2}
\begin{tabular}{l|c|c|c|c}
\hline
K         & 50    & 100    & 200    & 300     \\ \hline
LOF~\citep{breunig2000lof}         & 0.047& 0.084 & 0.120 & 0.158\\ \hline
Radar~\citep{li2017radar}      & 0.082& 0.156 & 0.282 & 0.336\\ \hline
ANOMALOUS~\citep{peng2018anomalous}   & 0.087& 0.158 & 0.289 & 0.340\\ \hline
DOMINANT~\citep{ding2019deep}   & 0.084& 0.162 & 0.304 & 0.396\\ \hline
MADAN~\citep{gutierrez2019multi}  & 0.078& 0.150 & 0.306 & 0.356\\ \hline
ResGCN (Our Model)   & \textbf{0.088} & \textbf{0.187} & \textbf{0.393} & \textbf{0.458} \\ \hline
\end{tabular}
\end{table*}

\begin{table*}
\centering
\caption{Performance of different anomaly detection methods w.r.t. recall@K on ACM. The bold indicates the best performance of all the methods.}
\label{tb:reck:3}
\begin{tabular}{l|c|c|c|c}
\hline
K         & 50    & 100    & 200    & 300     \\ \hline
LOF~\citep{breunig2000lof}       &0.005&0.010 &0.015 & 0.018\\ \hline
Radar~\citep{li2017radar}   &  0.047&0.097 &0.173 & 0.215\\ \hline
ANOMALOUS~\citep{peng2018anomalous} & 0.050&0.095 &0.170 & 0.205\\ \hline
DOMINANT~\citep{ding2019deep} & 0.052&0.098 &0.180 & 0.248\\ \hline
MADAN~\citep{gutierrez2019multi} &0.052&0.086 &0.210 & 0.225\\ \hline
ResGCN (Our Model)  &  \textbf{0.079} & \textbf{0.148} & \textbf{0.235} & \textbf{0.309}\\ \hline
\end{tabular}
\end{table*}

\subsection{Ranking Strategy Analysis}
\begin{figure*}
\centering
    \begin{subfigure}[b]{0.325\textwidth}
        \centering
        \includegraphics[width=\textwidth]{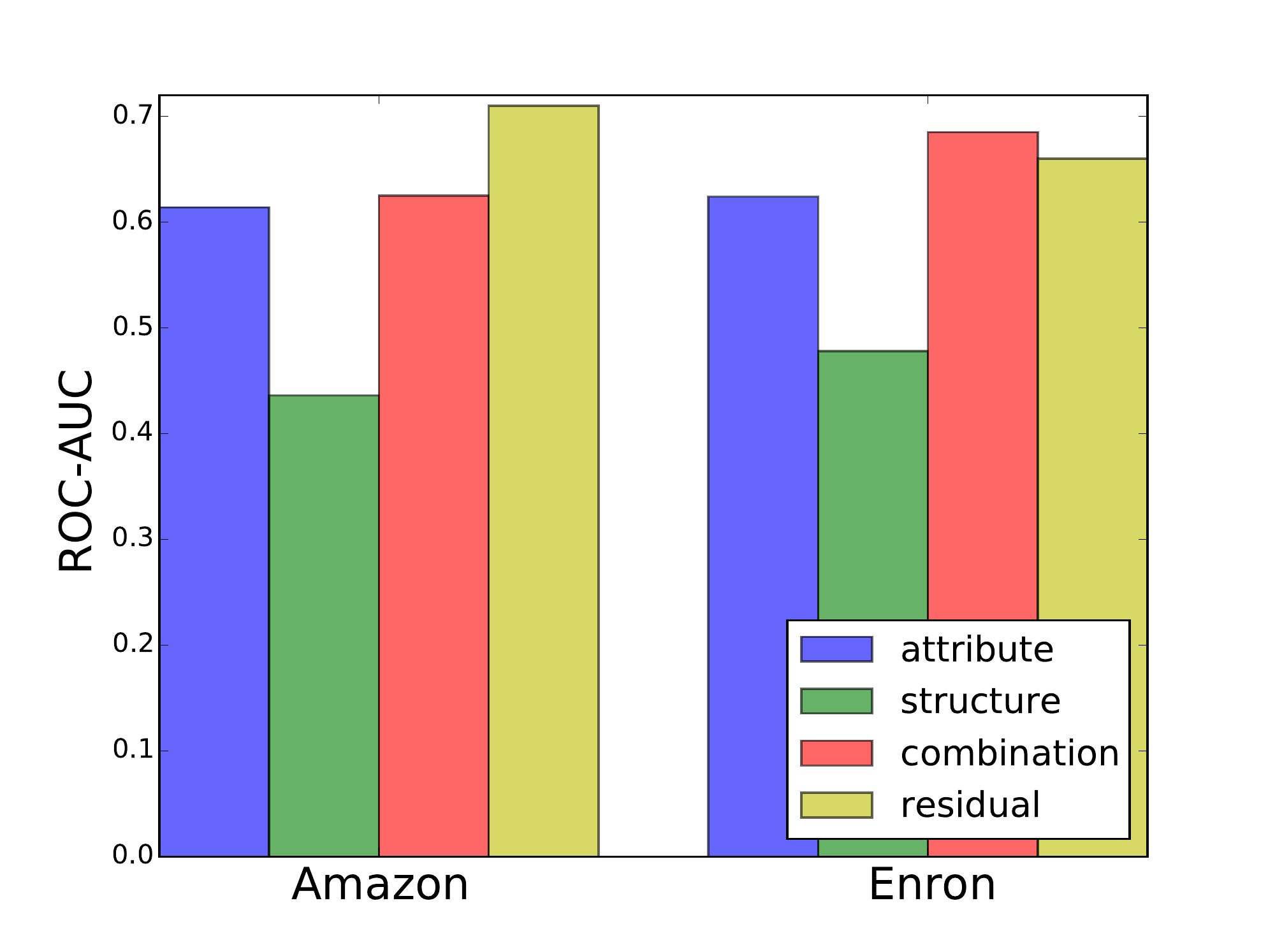}
        \caption[]%
        {{ROC-AUC}}    
        \label{fig:a}
    \end{subfigure}
    \hfill
    \begin{subfigure}[b]{0.325\textwidth}  
        \centering 
        \includegraphics[width=\textwidth]{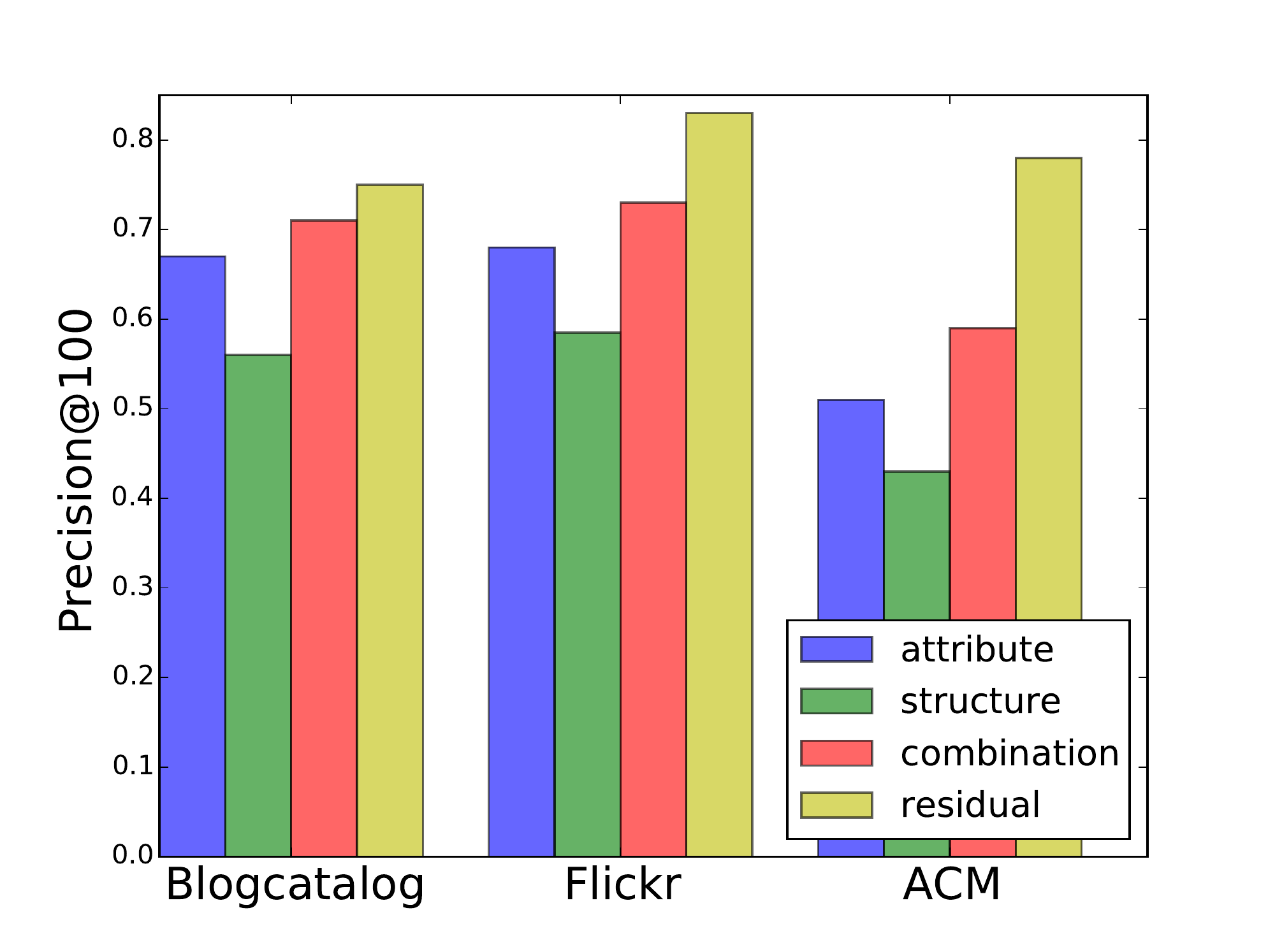}
        \caption[]%
        {{Precision@100}}    
        \label{fig:b}
    \end{subfigure}
    \hfill
    \begin{subfigure}[b]{0.325\textwidth}  
        \centering 
        \includegraphics[width=\textwidth]{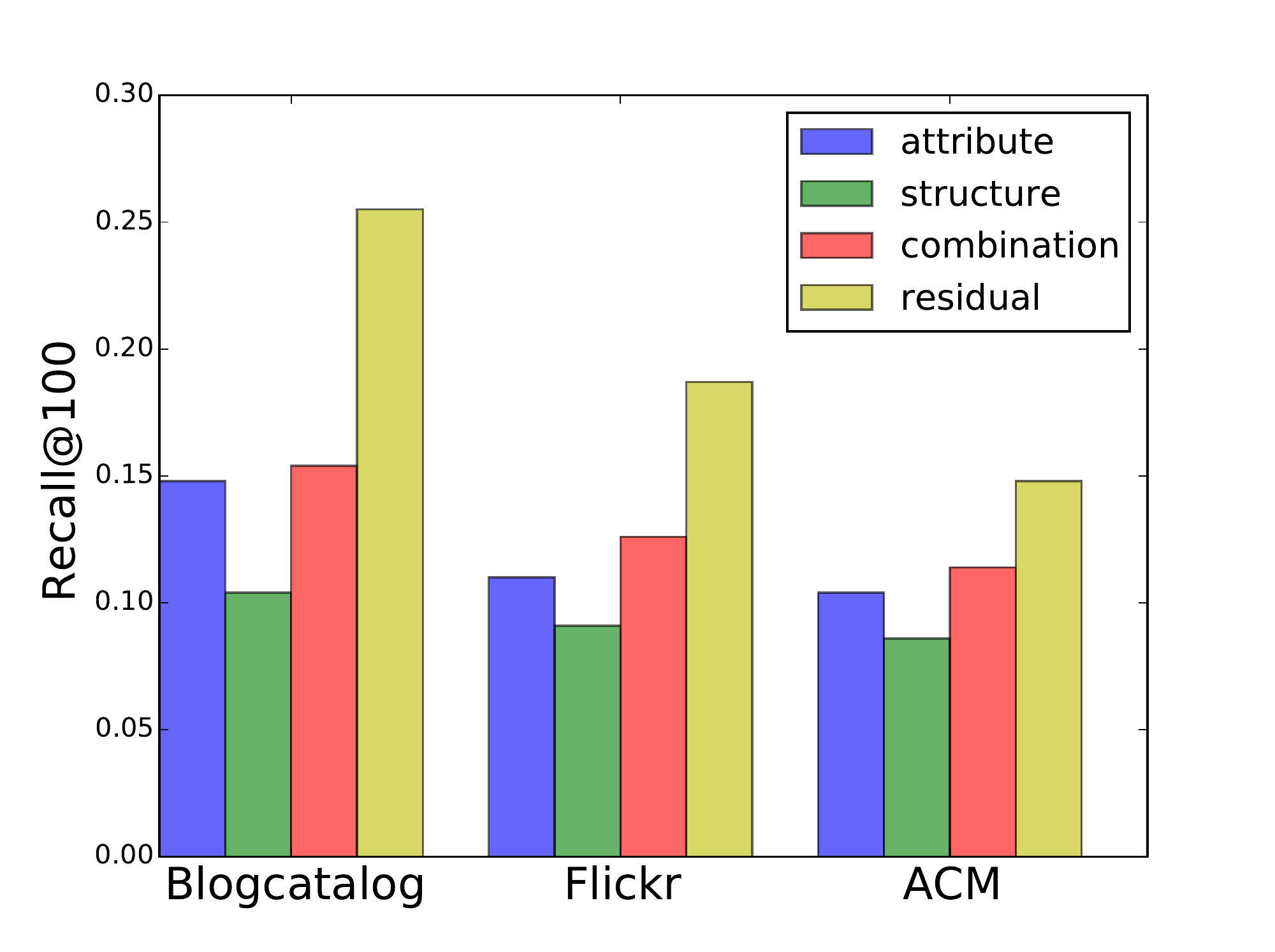}
        \caption[]%
        {{Recall@100}}    
        \label{fig:c}
    \end{subfigure}
    \caption[]
    {Comparison of different ranking strategies based on structure reconstruction, attribute reconstruction, combination of structures and attributes and residual information for anomaly detection: (a) ROC-AUC on Amazon and Enron, (b) Precision@100 on BlogCatalog, Flickr and ACM, and (c) Recall@100 on BlogCatalog, Flickr and ACM.} 
    \label{fig:rank}
\end{figure*}

One of the advantages of our proposed ResGCN is the deep residual modeling to capture the anomalous information. Therefore, different from DOMINANT~\citep{ding2019deep} which ranks the weighted combination of attribute and structure reconstruction errors to select the anomalous nodes, we rank the residual information for anomaly detection. In this section, we compare different ranking strategies for anomaly detection: (1) ranking attribute reconstruction error, (2) ranking structure reconstruction error, (3) ranking the weighted combination of attribute and structure reconstruction errors, and (4) ranking the residual matrix. The first three strategies have been used in~\citep{ding2019deep} and the last one has been used in Radar~\citep{li2017radar}. The results of anomaly detection w.r.t. ROC-AUC on Amazon and Precision@100 and Recall@100 on BlogCatalog are shown in Figure~\ref{fig:rank}. 

From the results, it can be observed that:
\begin{itemize}
    \item ranking the residual matrix outperforms other ranking strategies on all the data w.r.t. different evaluation metrics except on Enron data. It demonstrates the effectiveness of residual modeling in ResGCN for anomaly detection.
    \item By combining attribute and structure reconstruction errors, better detection performance can be achieved. This result indicates that both attributes and structures contain some useful information to detect anomalies.
    \item An interesting observation is that attributes play a more important role in detecting anomalies than structures as ranking attribute reconstruction errors performs better than structure construction errors.
\end{itemize}

\subsection{Parameter Analysis}
\begin{figure*}
\centering
    \begin{subfigure}[b]{0.32\textwidth}
        \centering
        \includegraphics[width=\textwidth]{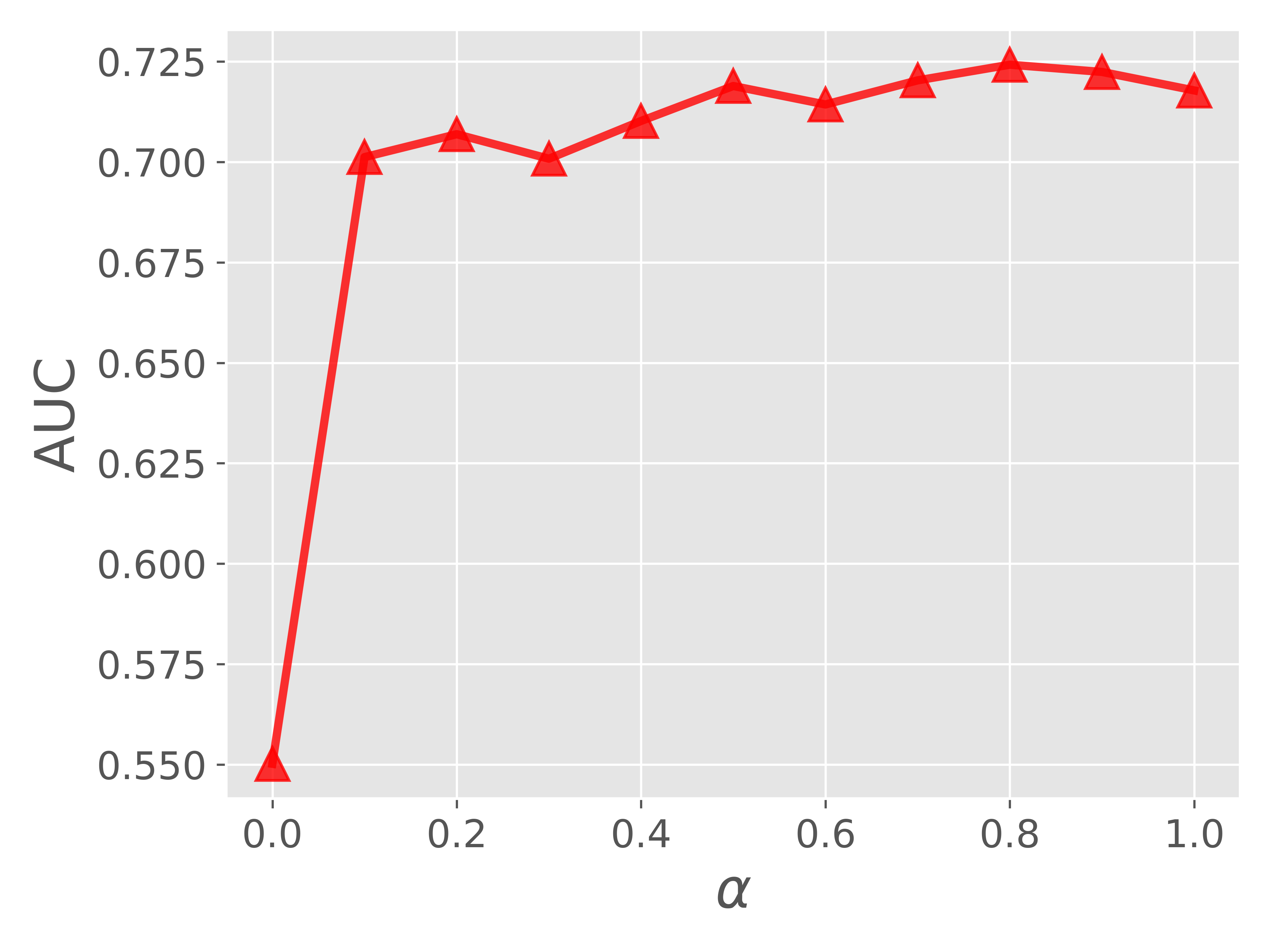}
        \caption[]%
        {{ROC-AUC on Amazon}}    
        \label{fig:a:a}
    \end{subfigure}
    \hfill
    \begin{subfigure}[b]{0.32\textwidth}  
        \centering 
        \includegraphics[width=\textwidth]{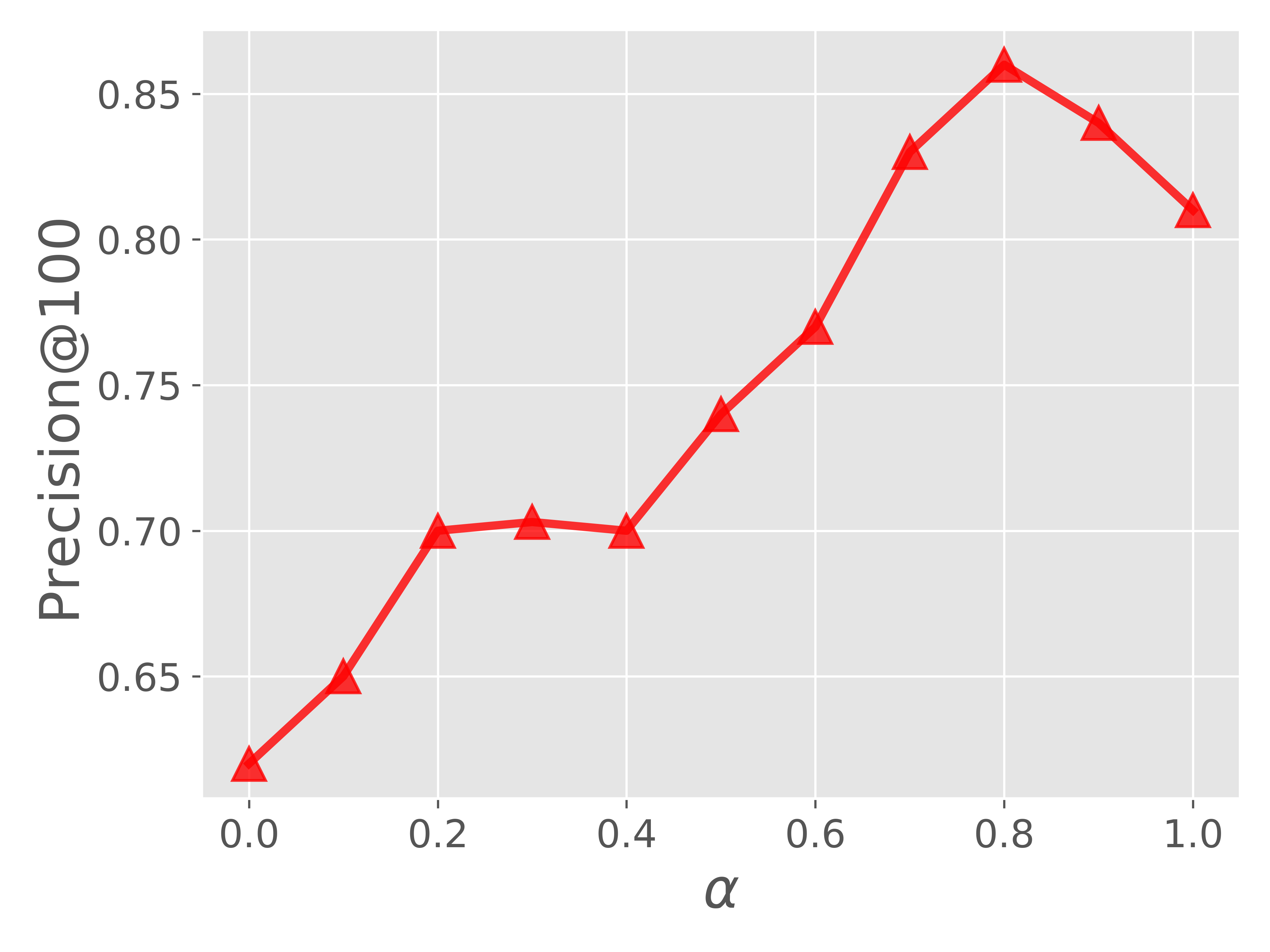}
        \caption[]%
        {{P@100 on BlogCatalog}}    
        \label{fig:a:b}
    \end{subfigure}
    \hfill
    \begin{subfigure}[b]{0.32\textwidth}  
        \centering 
        \includegraphics[width=\textwidth]{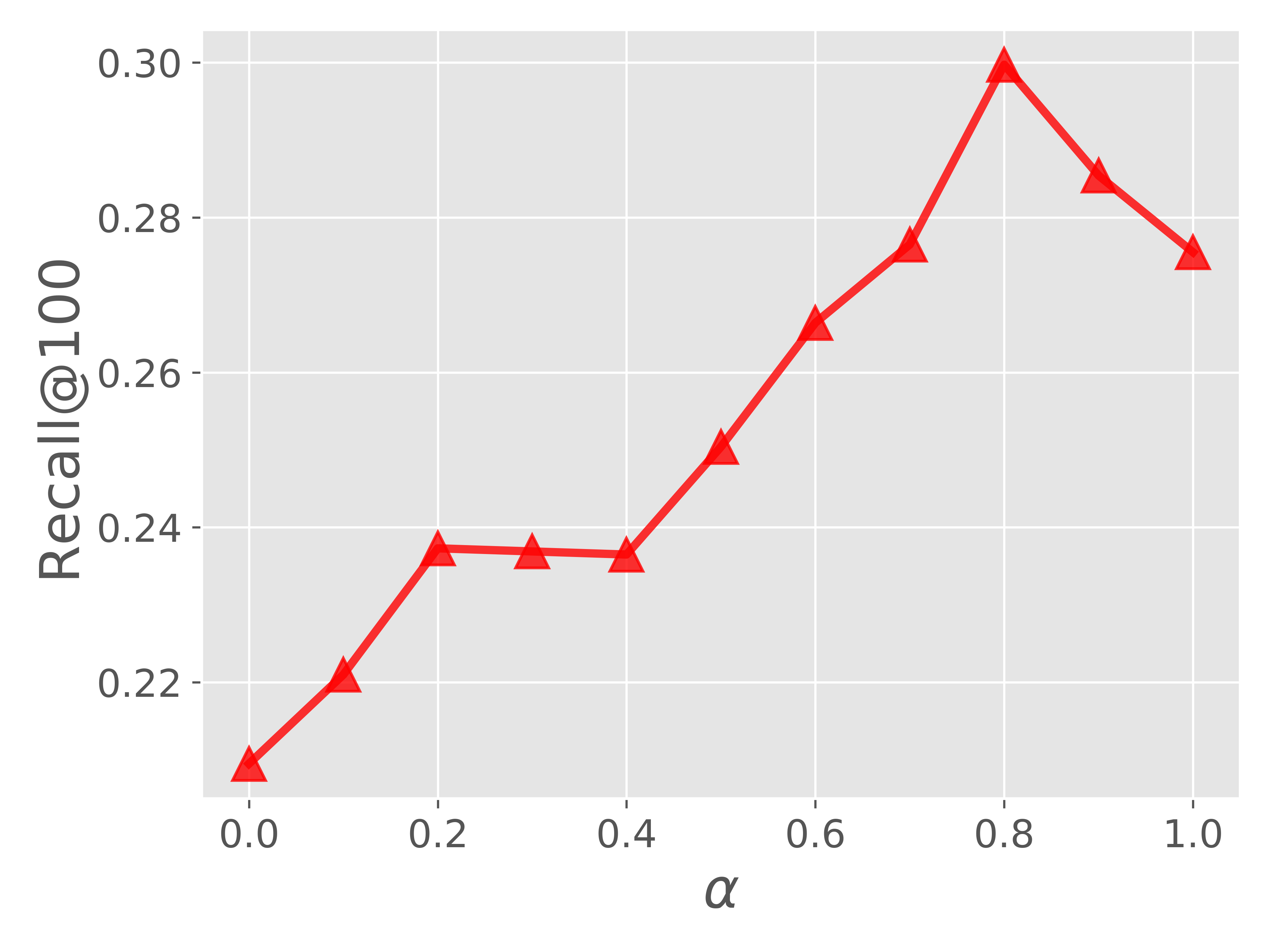}
        \caption[]%
        {{R@100 on BlogCatalog}}    
        \label{fig:a:c}
    \end{subfigure}
    \caption[]
    {Influence of the trade-off parameter $\alpha$ for structure and attribute reconstruction errors (ranging from 0.0 to 1.0): (a) ROC-AUC on Amazon, (b) Precision@100 on BlogCatalog, and (c) Recall@100 on BlogCatalog.} 
    \label{fig:alpha}
\end{figure*}

\begin{figure*}
\centering
    \begin{subfigure}[b]{0.32\textwidth}
        \centering
        \includegraphics[width=\textwidth]{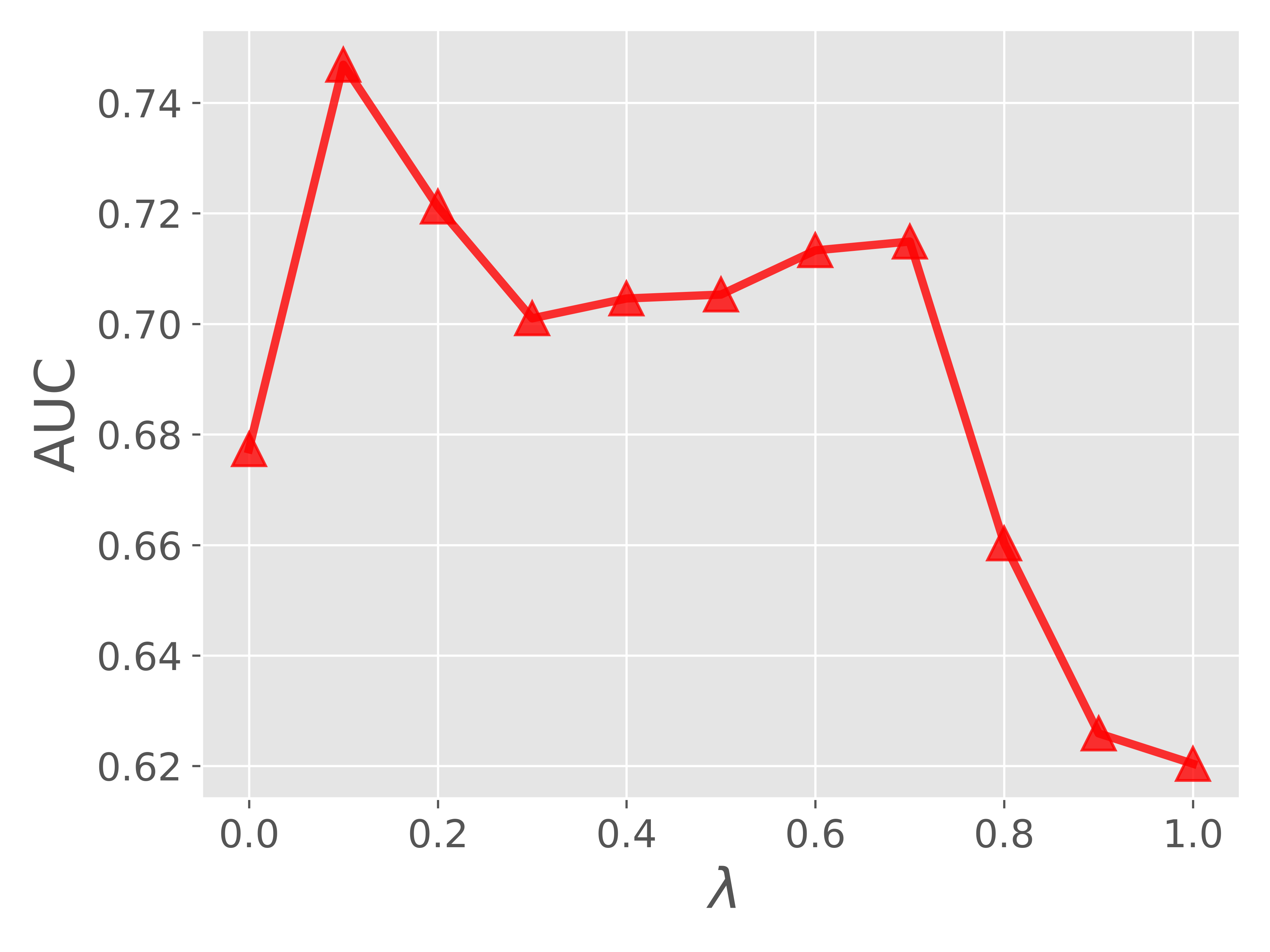}
        \caption[]%
        {{ROC-AUC on Amazon}}    
        \label{fig:l:a}
    \end{subfigure}
    \hfill
    \begin{subfigure}[b]{0.32\textwidth}  
        \centering 
        \includegraphics[width=\textwidth]{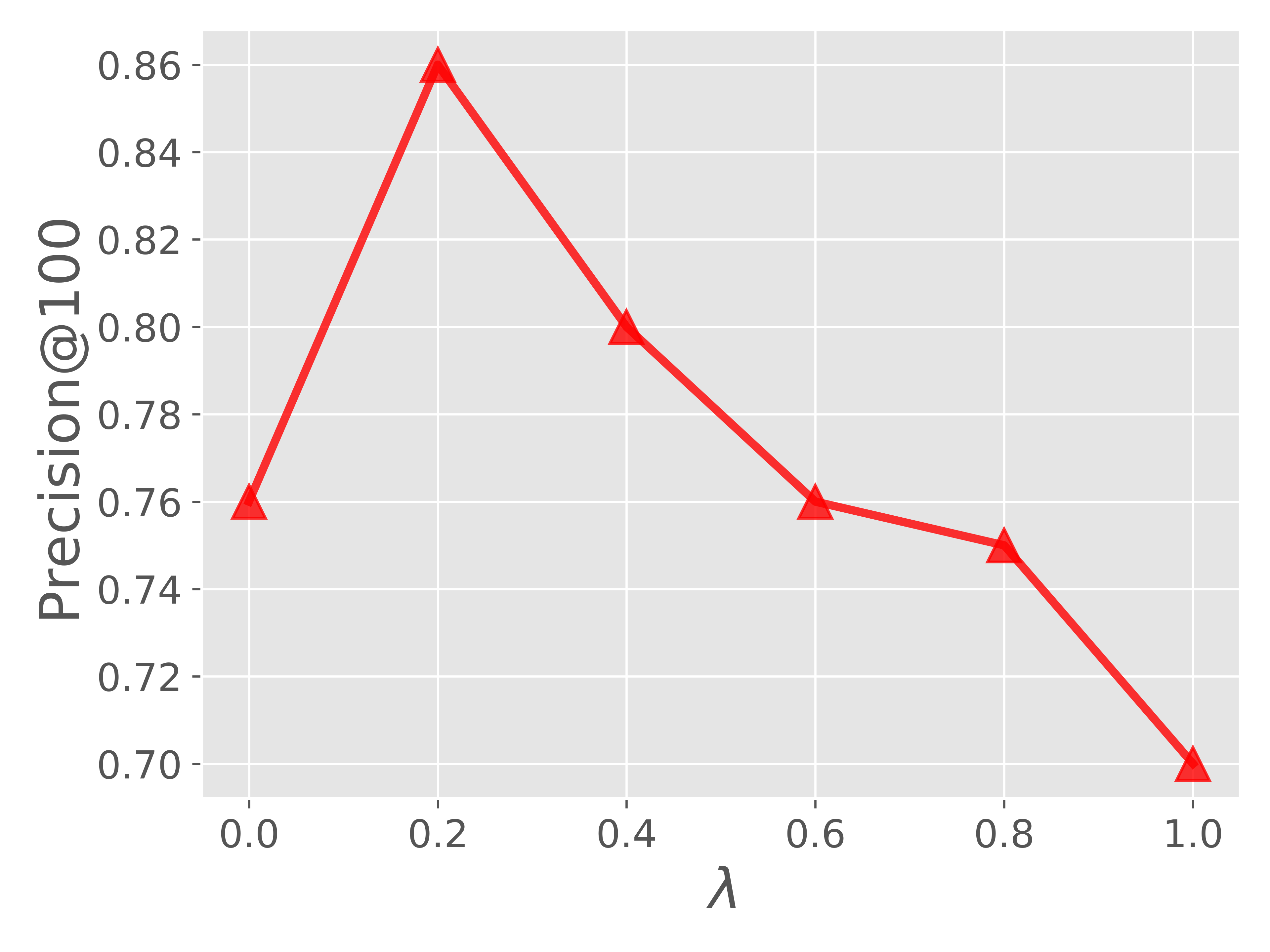}
        \caption[]%
        {{P@100 on BlogCatalog}}    
        \label{fig:l:b}
    \end{subfigure}
    \hfill
    \begin{subfigure}[b]{0.32\textwidth}  
        \centering 
        \includegraphics[width=\textwidth]{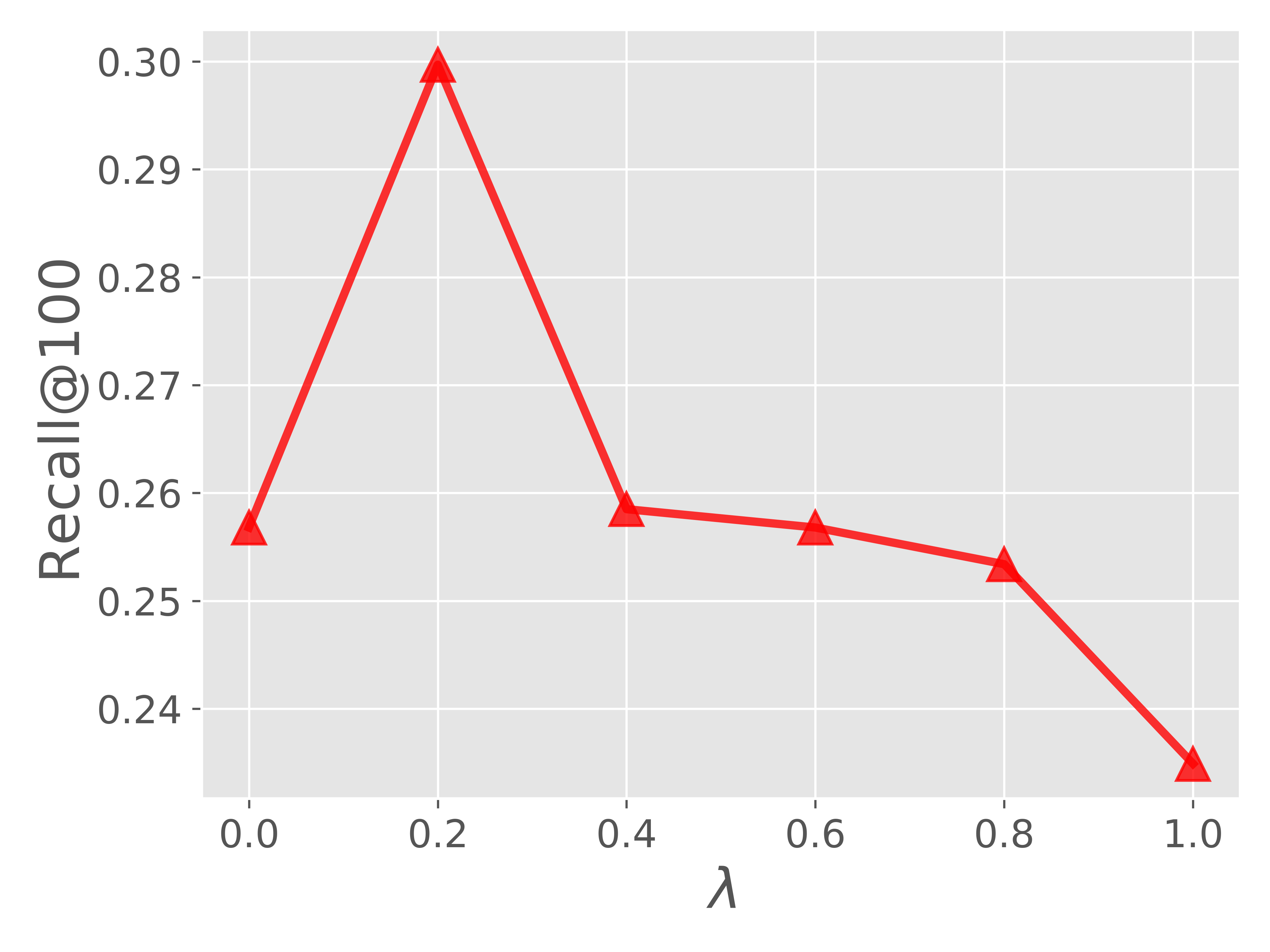}
        \caption[]%
        {{R@100 on BlogCatalog}}    
        \label{fig:l:c}
    \end{subfigure}
    \caption[]
    {Influence of the residual parameter $\lambda$ for loss function (ranging from 0.0 to 1.0): (a) ROC-AUC on Amazon, (b) Precision@100 on BlogCatalog, and (c) Recall@100 on BlogCatalog.} 
    \label{fig:lambda}
\end{figure*}

There are different parameters in our proposed ResGCN model. Among them, there are two specific and important ones: (1) the trade-off parameter $\alpha$ for structure and attribute reconstruction errors and (2) the residual parameter $\lambda$ in the loss function in Eq (\ref{recons}). In this experiment, we investigate the impact of these two parameters separately. Specifically, we test the anomaly detection performance by ranging $\alpha$ and $\lambda$ from 0.0 to 1.0 on Amazon and BlogCatalog data respectively. The results for $\alpha$ and $\lambda$ are shown in Figure~\ref{fig:alpha} and Figure~\ref{fig:lambda} respectively.

From the results, it can be observed that:
\begin{itemize}
    \item The influence of $\alpha$ shows different trends on different networks. For Amazon, the performance becomes much better when $\alpha\ge 0.1$. For BlogCatalog, larger $\alpha$ achieves better performance. The commonness is that it achieves the best performance when $\alpha=0.8$ on both networks.
    \item The impact of $\lambda$ is similar on different networks, i.e., both Amazon and BlogCatalog prefer smaller $alpha$. Empirically the best detection performance can be achieved when $\lambda=0.1$ on Amazon and $\lambda=0.2$ on BlogCatalog.
\end{itemize}

\section{Related Work}
\label{rw}

Anomaly detection is one of the most important research questions in data mining and machine learning. There are different anomalies in different types of data, e.g., text~\citep{kannan2017outlier,ruff2019self}, network~\citep{bhuyan2013network} and temporal data~\citep{gupta2013outlier}. Earlier studies of anomaly detection on graphs mainly focused on structural anomalies, e.g.,~\citep{noble2003graph} and~\citep{eberle2007discovering}. However, compared to anomaly detection approaches on plain networks, anomaly detection on attributed networks is more challenging because both structures and attributes should be taken into consideration. In this section, we concentrate on the related work of anomaly detection on attributed networks.

Real-world networks often come with auxiliary attribute information, so recent years have witnessed an increasingly amount of efforts in detecting anomalies on attributed networks. Existing anomaly detection approaches on attributed networks can be categorized into several different types~\citep{ding2019deep}: community analysis, subspace selection,residual analysis and deep learning methods.

CODA~\citep{gao2010community} focuses on community anomalies by simultaneously finding communities as well as spotting anomalies using a unified probabilistic model. AMEN~\citep{perozzi2016scalable} uses both attribute and network structure information to detect anomalous neighborhoods. Radar~\citep{li2017radar} detects anomalies whose behaviors are singularly different from the majority by characterizing the residuals of attribute information and its coherence with network information. ANOMALOUS~\citep{peng2018anomalous} is a joint anomaly detection framework to optimize attribute selection and anomaly detection using CUR decomposition of matrix and residual analysis on attributed networks. DOMINANT~\citep{ding2019deep} utilizes GCN to compress the input attributed network to succinct low-dimensional embedding representations and then reconstruct both the topological structure and nodal attributes with these representations. MADAN~\citep{gutierrez2019multi} is a multi-scale anomaly detection method. It uses the heat kernel as filtering operator to exploit the link with the Markov stability to find the context for outlier nodes at all relevant scales of the network. For traditional anomaly detection methods on graphs, interested readers are referred to~\citep{akoglu2015graph} for detailed discussion.

With the popularity of network embedding techniques, which assigns nodes in a network to low-dimensional representations and these representations can effectively preserve the network structure~\citep{cui2018survey}, learning anomaly aware network representations also attracts huge attentions. Recently, there are several studies taking both problems into consideration to learn anomaly aware network embedding in attributed networks~\citep{liang2018semi,zhou2018sparc,bandyopadhyay2019outlier,li2019specae,bandyopadhyay2020outlier}. SEANO~\citep{liang2018semi} is a semi-supervised network embedding approach which learns a low-dimensional vector representation that systematically captures the topological proximity, attribute affinity and label similarity of nodes. SPARC~\citep{zhou2018sparc} is a self-paced framework for anomaly detection which gradually learns the rare category oriented network representation. ONE~\citep{bandyopadhyay2019outlier} jointly align and optimize the structures and attributes to generate robust network embeddings by minimizing the effects of outlier nodes. DONE and AdONE~\citep{bandyopadhyay2020outlier} use two parallel autoencoders for link structure and attributes of the nodes respectively. By exploring the reconstruction errors for structures and attributes, the proposed methods can learn embedding and detect anomalies. Another related embedding methods aim to capture the uncertainties of learned representations, such as and DVNE~\citep{zhu2018deep} \textit{struc2gauss}~\citep{pei2020struc2gauss}, where each node is mapped to a Gaussian distribution and the variance can capture the uncertainties. Intuitively, nodes with higher uncertainties are more likely to be anomalous.

Another related work is graph convolutional networks (GCNs). The original GCN~\citep{kipf2016semi} have been proposed to learn node representations by passing and aggregating messages between neighboring nodes. Different variants extend GCN have been proposed, e.g., introducing attention~\citep{velickovic2017graph}, adding residual and jumping connections~\citep{xu2018representation} and disentangling node representations~\citep{ma2019disentangled}.

\section{Conclusions}
\label{conc}
In this paper, we proposed a novel graph convolutional network (GCN) with attention mechanism, ResGCN, on the problem of anomaly detection on attributed networks. ResGCN can effectively address the limitations of previous approaches. On one hand, as GCN handles the high-order node interactions with multiple layers of nonlinear transformations, ResGCN can capture the sparsity and nonlinearity of networks. On the other hand, the attention mechanism based on the explicit deep residual analysis can prevent anomalous nodes from propagating the abnormal information in the message passing process of GCN. Furthermore, ranking the residual information is employed to detect anomalies. The experimental results demonstrate the effectiveness of our proposed ResGCN model compared to state-of-the-art methods. In the future, we would like to investigate the extension of our model in dynamic and streaming networks.

\bibliographystyle{spbasic} 
\bibliography{ref.bib} 

\end{document}